\documentclass[10pt,journal,compsoc]{IEEEtran}
\usepackage{graphicx}
\usepackage{hyperref}
\usepackage{amsmath}
\usepackage{multirow}
\usepackage[center]{caption}
\usepackage{amssymb}
\usepackage{placeins}

\ifCLASSOPTIONcompsoc
  \usepackage[nocompress]{cite}
\else
  \usepackage{cite}
\fi
\ifCLASSINFOpdf
\else
\fi
\begin{document}
%


\title{Audio-Visual Fusion for Emotion Recognition in the Valence-Arousal Space Using Joint Cross-Attention}
%
%
%
%

\author{R~Gnana~Praveen,~\IEEEmembership{}
        Patrick~Cardinal,~\IEEEmembership{Member,~IEEE,}
        and~Eric~Granger,~\IEEEmembership{Member,~IEEE}
\IEEEcompsocitemizethanks{\IEEEcompsocthanksitem The authors are with the Laboratoire d'imagerie, de vision et d'intelligence artificielle (LIVIA), Dept. of Systems Engineering, Ecole de technologie supérieure, Montreal, Canada.\protect\\
E-mail: gnanapraveen.rajasekar.1@ens.etsmtl.ca (R. G. Praveen),  patrick.cardinal@etsmtl.ca (P. Cardinal), eric.granger@etsmtl.ca (E. Granger)) 
}
}

%
%

\markboth{Journal of \LaTeX\ Class Files,~Vol.~14, No.~8, August~2015}%
{Shell \MakeLowercase{\textit{et al.}}: Bare Demo of IEEEtran.cls for Biometrics Council Journals}
%



\IEEEtitleabstractindextext{%
\begin{abstract}
Automatic emotion recognition (ER) has recently gained lot of interest due to its potential in many real-world applications. In this context, multimodal approaches have been shown to improve performance (over unimodal approaches) by combining diverse and complementary sources of information, providing some robustness to noisy and missing modalities. In this paper, we focus on dimensional ER based on the fusion of facial and vocal modalities extracted from videos, where complementary audio-visual (A-V) relationships are explored to predict an individual's emotional states in valence-arousal space. Most state-of-the-art fusion techniques rely on recurrent networks or conventional attention mechanisms that do not effectively leverage the complementary nature of A-V modalities. To address this problem, we introduce a joint cross-attentional model for A-V fusion that extracts the salient features across A-V modalities, that allows to effectively leverage the inter-modal relationships, while retaining the intra-modal relationships. In particular, it computes the cross-attention weights based on correlation between the joint feature representation and that of the individual modalities. By deploying the joint A-V feature representation into the cross-attention module, it helps to simultaneously leverage both the intra and inter modal relationships, thereby significantly improving the performance of the system over the vanilla cross-attention module. The effectiveness of our proposed approach is validated experimentally on challenging videos from the RECOLA and AffWild2 datasets. Results indicate that our joint cross-attentional A-V fusion model provides a cost-effective solution that can outperform state-of-the-art approaches, even when the modalities are noisy or absent. Code is available at \url{ https://github.com/praveena2j/Joint-Cross-Attention-for-Audio-Visual-Fusion}. 
\end{abstract}

}


\maketitle

\IEEEdisplaynontitleabstractindextext

%
\IEEEpeerreviewmaketitle

\IEEEraisesectionheading{\section{Introduction}\label{sec:introduction}}

%
%
%
%
\IEEEPARstart{A}{utomatic} recognition and analysis of human emotions has drawn much attention over the past few decades. It has been extensively researched in various fields such as neuroscience, psychology, cognitive science and computer science, leading to the advancement of a wide range of applications in various fields, such as health care (e.g., assessment of anger, fatigue, depression and pain), robotics (human-machine interaction), driver assistance (assessment of driver's state), etc \cite{Kolakowska2014}.  Emotion recognition (ER)  is a challenging problem since the expressions linked to human emotions are extremely diverse in nature across individuals and cultures. Ekman conducted cross-cultural study on human emotions, and categorized the basic emotions into six categories -- anger, disgust, fear, happy, sad, and surprise \cite{Ekman}. Subsequently, contempt has been added to these six basic emotions \cite{Matsumoto}. The categorical model of ER has been explored extensively in the field of affective computing due to its simplicity and universality \cite{Anagnostopoulos}. 
\begin{figure}[!t]
\centering
\includegraphics[width=0.45\textwidth]{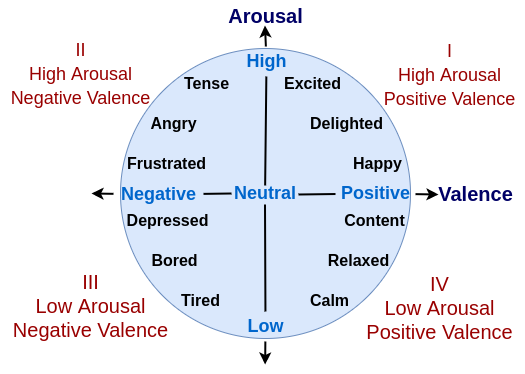}
\caption{The valence-arousal space. Valence denotes the range of emotions from being very sad (negative) to very happy (positive) and arousal reflects the energy or intensity of emotions from very passive to very active.}
\label{fig:VA}
\end{figure}
Recently, real-world applications have driven a shift of affective computing research from laboratory-controlled environments to more realistic natural settings. This shift has further led to the analysis of wide range of subtle, continuous emotional and health states that are elicited in real-world settings. Conventionally, the estimation of continuous ER states are formulated as the dimensional ER problem, where complex human emotions can be represented in a dimensional valence-arousal space. Figure \ref{fig:VA} illustrates the use of a two-dimensional space to represent emotional states, where valence and arousal are employed as dimensional axes \cite{Schlosberg}. 
Valence reflects the wide range of emotions in the dimension of pleasantness from being negative (sad) to positive (happy), whereas arousal spans the range of intensities from passive (sleepiness) to active (high excitement) \cite{5740839}. Recognizing such fine-grained emotional states is beneficial in  various applications, such as assessing driver fatigue, estimating the level of depression or pain in health-care, assessing customer engagement in marketing, etc. Given the growing need for continuous ER in real-world applications, this paper focuses on dimensional ER in the valence-arousal space.

Human emotions can be conveyed through various modalities such as face, voice, text and physiology (electroencephalogram, electrocardiogram, etc.), which typically carry complementary information among them. Although human emotions can be expressed through various modalities, vocal and facial modalities are the predominant contact-free channels, which carries complementary information \cite{5551170}. 
Audio-visual (A-V) fusion has also been widely explored for various applications including identity verification \cite{786997}, event localization \cite{9423042}, action recognition \cite{lee2021crossattentional}, etc. Efficiently leveraging the complementary nature of A-V relationships captured in videos can play a crucial role in improving the performance of multimodal systems over unimodal systems \cite{8683477}. 
Techniques for multimodal fusion can be broadly categorized as model-agnostic or model-based \cite{8269806}. In model-based approaches, fusion is performed using specialized models to cope with the diverse information in multimodal data. 
Depending on the type of model used for fusion, these techniques are typically classified further as kernel methods, graphical models, or neural networks \cite{8269806}. Unlike model based fusion, model agnostic fusion can be achieved using almost any uni-modal classifier or regressor. They do not rely on any specialized model for fusion. Most of the existing fusion models belongs to this category, where fusion is often performed by concatenating the features or individual modal predictions.
Model agnostic approaches can be further classified as three major strategies: decision-, feature-, hybrid-level fusion \cite{wu_lin_wei_2014}. In decision-level fusion (late fusion), multiple modalities are trained end-to-end independently, and then the predictions obtained from the individual modalities are fused to obtain the final predictions. Although decision-level fusion is easy to implement, and requires less training, it neglects the interactions across the individual modalities, thereby resulting in limited improvement over uni-modal approaches. Conventionally, feature-level fusion (early fusion) is achieved by concatenating the features of A-V modalities immediately after they are extracted, which is further used for predicting the final outputs. Hybrid fusion takes advantage of both decision-level and feature-level fusion by combining outputs from both feature-level fusion and decision level fusion. Though feature level fusion is conventionally done by aggregating or concatenating the features immediately after they are extracted, it can also be performed by learning the interactions between the modalities for better feature representations before concatenating the features \cite{9607460, 9667055}. In this work, we explore feature-level fusion based on joint cross-attention, where the A and V features extracted from videos are further modeled using a joint cross-attention model prior to concatenation. 

Deep learning (DL) models provide state-of-the-art performance in many V recognition applications, such as image classification, object detection, action recognition, etc. Inspired by their performance, several ER approaches have been proposed for video-based dimensional ER using CNNs to obtain the deep features, and a recurrent neural network to capture the temporal dynamics \cite{cite6,cite7}. Deep models have also been widely explored for vocal emotion recognition, typically using spectrograms with 2D-CNNs \cite{cite6, 9607711}, or raw wave forms with 1D-CNNs \cite{cite7}. In most of the existing approaches \cite{cite7, cite8} for dimensional ER, A-V fusion is performed by concatenating the deep features extracted from individual facial and vocal modalities, and fed to LSTM for predicting valence and arousal. Although LSTM based fusion models can improve the system performance by leveraging the intra-modal relationships, it does not effectively capture the inter-modal relationships across the individual modalities. We therefore investigate the prospect of extracting more comprehensive salient features that can effectively exploit the complementary relationships across the A and V modalities.

Attention mechanisms have recently gained much interest in the areas of computer vision and machine learning as they allow extracting task relevant features, thereby improving system performance. This has been extensively explored for various applications, such as event/action recognition \cite{9157522}, ER \cite{Jiyoung}, etc. Most of the existing attention based approaches for dimensional ER explore the intra-modal relationships \cite{Jiyoung}. Although a few approaches \cite{srini_2021_SLT, cite8} attempt to capture the cross-modal relationships using cross-attention based on transformers, they fail to effectively leverage the complementary relationship of A-V modalities. Indeed, their computation of attention weights does not consider the correlation across the A and V features. 

A preliminary version of the cross-attentional (CA) A-V fusion model for dimensional ER was presented in our previous work \cite{9667055}. In this work, we further extend our previous work, where a joint A-V feature representation is deployed in the CA model in order to jointly capture both intra and inter modal relationships. In this previous paper \cite{9667055}, the attention weights are computed based on the correlation across A and V modalities, which depends only on intermodal relationships. Instead of using individual feature representations across the modalities to generate the attention weights, we introduce joint A-V feature representations to capture the relationships within the same modality as well as other modality, thereby leveraging both inter- and intra-modal relationships to obtain the attention weights. Using the joint feature representation drastically reduces the heterogeneity across the A and V features, which further helps to provide robust A-V feature representations. Specifically, we obtain the cross-correlation matrix across the deep joint feature representation and features of individual modalities to obtain the attention weights for the A and V modalities. Therefore, the attention weights of each modality is obtained not only using the features of itself but also from the other modality, resulting in more informative features. Besides providing improved performance over individual modalities, a benefit of our joint A-V representation is its ability to perform well even when a modality is noisy or absent. Finally, we have also explored the impact on JCA performance of feature-level fusion, where multiple divers backbones are combined for the A and V modalities. 

The main contributions of the paper are: 
(1) A joint cross-attentional (JCA) model for A-V fusion is introduced to effectively exploit the complementary relationship across modalities for dimensional ER in valence-arousal space. Contrary to the prior approaches, the proposed model simultaneously leverages both intra and inter modal relationships to effectively capture the complementary relationships. (2) By deploying the joint feature representation it also helps to reduce the heterogeneity across A and V features, thereby resulting in robust AV feature representations.  (3) An extensive set of experiments on the challenging RECOLA and Affwild2 datasets indicate that our proposed JCA fusion model can outperform related state-of-the-art fusion models for dimensional ER. Our visual interpretation of the fusion process shows that JCA can efficiently leverages the complementary intermodal relationships, while retaining the intramodal relationships. 

The rest of this paper is organized as follows. Section II provides a critical analysis of the relevant literature on dimensional ER, and attention models for A-V fusion. Section III describes the proposed JCA A-V fusion model in detail. Section IV presents the experimental methodology for the backbones of the individual modalities, and the experimental settings used in our fusion model. Finally, the results obtained with the proposed approach with RECOLA and Affwild2 datasets are presented and discussed in Section V.

\section{Related Work}
\subsection{A-V Fusion for Dimensional Emotion Recognition:}
One of the early approaches using DL models for A-V fusion based dimensional ER was proposed by Tzirakis et al. \cite{cite7}, where A and V features are obtained using ResNet50 and 1D-CNN respectively. The obtained features are then concatenated and fed to Long short-term memory model (LSTM) for the prediction of valence and arousal. Juan et al. \cite{8914655} investigated an empirical study of fine-tuning pretrained CNN models by freezing various convolutional layers. Schonevald et al. \cite{cite6} explored knowledge distillation using teacher-student model for V modality and CNN model for A modality using spectrograms. The deep feature representations are combined using model-based fusion strategy, where RNNs are used to capture the temporal dynamics. Inspired by the deep auto-encoders, Nguyen et al. \cite{9374787} investigated the prospect of how to simultaneously learn compact representative features from A and V modalities using deep auto-encoders. They have proposed a deep model of two-stream auto-encoders and LSTM for efficiently integrating V and A streams for dimensional ER. 

Deng et al \cite{9607738} proposed iterative self distillation method for modeling the uncertainties in the labels in a multi-task framework. They have trained a model with multiple task labels, which is further used to distill iteratively to several student models. They have shown that iterative distillation significantly improves the performance of the system. Kuhnke et al. \cite{9320301} proposed two stream A-V network, where V features are extracted from R(2plus1)D model \cite{8578773} pretrained from action recognition dataset and A features are obtained from Resnet18 model \cite{7780459}. The obtained features are further concatenated for final prediction of valence and arousal. Wang et al \cite{9607711} further improved their approach \cite{9320301} by introducing teacher-student model in a semi-supervised learning framework. The teacher model is trained on the available labels, which is further used to obtain pseudo labels for unlabeled data. The pseudo labels are finally used to train the student model, which is used for final prediction. Though the above mentioned approaches have shown significant improvement for dimensional ER, they fail to effectively capture the inter-modal relationships and relevant salient features specific to the task. Therefore, we have focused on capturing the comprehensive features in a complementary fashion using attention mechanisms.  

\subsection{Attention Models for A-V Fusion:}
Attention mechanisms are widely used in the context of multimodal fusion with various modalities such as A and text \cite{Lee2020,N.2020}, V and text \cite{8578827, Wei_2020_CVPR}, etc. Zhao et al. \cite{Zhao_Ma_Gu_Yang_Xing_Xu_Hu_Chai_Keutzer_2020} proposed an end-to-end architecture for emotion classification by integrating spatial, channel-wise and temporal attentions into V network and temporal attention into A network. Esam et al. \cite{9191019} explored attention to weigh the time windows of a video sequence to efficiently exploit the temporal interactions between the A-V modalities. They used transformer \cite{NIPS2017_3f5ee243} based encoders to obtain the attention weights through self attention for emotion classification. Lee et al. \cite{Jiyoung} proposed spatiotemporal attention for the V modality to focus on emotional salient parts using Convolutional LSTM (ConvLSTM) modules and a temporal attention network using deep networks for A modality. Then the attended features are concatenated and fed to the regression network for the prediction of valence and arousal. However, these approaches focused on modeling the intra-modal relationships and failed to effectively exploit the inter-modal relationship of the A-V modalities.

Wang et al. \cite{Wang_ICMI_2020} investigated the prospect of exploiting the implicit contextual information along with the A and V modalities. They have proposed an end-to-end architecture using cross-attention based on transformers for A-V group ER. Srinivas et al. \cite{srini_2021_SLT} also explored transformers with cross-modal attention for dimensional ER, where cross-attention is integrated along with self attention. Tzirakis et al. \cite{cite8} investigated various fusion strategies along with attention mechanisms for A-V fusion based dimensional ER. They have further explored self attention as well as cross-attention fusion based on transformers in order to enable the extracted features of different modalities to attend to each other. Although these approaches have explored cross-modal attention with transformers, they fail to leverage semantic relevance among the A-V features based on cross-correlation.

Zhang et al. \cite{9607460} investigated the prospect of improving the fusion performance over individual modalities and proposed leader-follower attentive fusion for dimensional ER. 
The obtained features are encoded and attention weights are obtained by combining the encoded A and V features. The attention weights are further attended on the V features and concatenated to the original V features for final prediction.  Zhang et al. \cite{9320215} proposed attentive fusion mechanism, where V features are obtained from 3D-CNNs and A features from spectrograms fed to 2D-CNN. The obtained A and V features are further re-weighted using weights, obtained from scoring functions based on the relevant information in the individual modalities. Wang et al. \cite{9197622} addressed the problem of multi-modal feature fusion along with frame alignment issues between A and V modalities using cross-attention for speech recognition. Luo et al. \cite{luo18_interspeech} investigated the potential of joint representation learning using Convolutional Recurrent Neural Networks (CRNN) for vocal ER. They have also shown that the impact of time interval significantly impacts the performance of the system. Hu et al \cite{8683898} proposed dense multi-modal fusion by densely integrating the representation at multiple shared layers to capture hierarchical correlations across the modalities. Vedran et al \cite{10.1145/2911996.2912064} proposed a cross-modal deep network architecture, where the weights of two deep networks are enforced to be symmetry, yielding joint representation in a common feature space. In this work, we have used simple joint representation of feature concatenation of A and V modalities in our JCA framework.

Unlike prior approaches, we advocate for a simple yet efficient JCA model based on joint modeling of intra and inter modal relationships between A and V modalities. 
Cross-attention has been successfully applied in several applications, such as weakly-supervised action localization \cite{lee2021crossattentional}, and few-shot classification \cite{NEURIPS2019_01894d6f}. The similar idea of exploiting the complementary relationships for better audiovisual fusion has also been explored for person verification \cite{8683477}, where an attention layer is used for the fusion of A and V modalities. In most of these cases, cross-attention has been applied across the individual modalities. However, we have explored joint attention between individual and combined AV-features. By deploying the joint AV feature representation, we can effectively capture the intra and inter-modal relationships simultaneously by allowing interactions across the modalities as well as within oneself.  
Recently, joint co-attention has been explored by Duan et al. \cite{9423042} in a recursive fashion for A-V event localization. They have shown that recursive training of joint co-attention yields more discriminant and robust feature representations for multimodal fusion. In this paper, joint (combined) A-V features are extracted through cross-attention (instead of co-attention) for dimensional ER. Specifically, the features of each modality attend to themselves, as well as those of the other modality, through cross-correlation of the concatenated  A-V features, and features of individual modalities. By effectively leveraging the joint modeling of intra- and inter-modal relationships, the proposed approach can significantly improve system performance. 


\section{Proposed Approach}

\subsection{Visual Network:} \label{visual NN}
Facial expressions from videos involve both appearance and temporal dynamics of video sequences. Efficient modeling of these spatial and temporal dynamics play a crucial role in extracting discriminant and robust features, which in-turn improves the overall system performance. State-of-the-art performance is typically achieved using 2D-CNN in combination with Recurrent Neural Networks (RNN) to capture the effective latent appearance representation, along with temporal dynamics \cite{7904596}. Several approaches have been explored for dimensional facial ER based on 2D-CNNs and LSTMs \cite{5740839}, \cite{WOLLMER2013153}. However, 3D-CNNs are found to be efficient in capturing the spatiotemporal dynamics in videos \cite{RAJASEKHAR2021104167}, and have also been explored for dimensional facial ER. For instance, in \cite{9320301}, they have shown that R3D \cite{8578773} pretrained on the Kinetics-400 action recognition dataset \cite{kinetics} has outperformed  conventional 2D-CNNs for dimensional ER on Affwild2 dataset. Inspired by the performance of 3D-CNNs, we consider Inflated 3D-CNN \cite{8099985}, to extract spatiotemporal features of the facial clips from a video sequence. Initially, proposed by Carreira et al. \cite{8099985} for action recognition, the Inflated 3D (I3D) CNN model can efficiently capture the spatiotemporal dynamics of the V modality while optimizing fewer parameters than that of conventional 3D-CNNs. I3D model is obtained by inflating the filters and pooling kernels of 2D ConvNet, expanding to 3D CNN. Therefore, it allows leveraging existing common pretrained 2D-CNNs, which are trained on large-scale image datasets for facial expressions, thereby improving the spatial discrimination for videos. Though I3D model has been primarily explored for action recognition, it has also been used for other applications in the field of affective computing, like in video-based pain localization \cite{9320216}, etc. In the proposed approach, we train the I3D model to extract spatio-temporal features for the facial  modality (see implementation details in Section \ref{Training details}).

\subsection{Audio Network:} \label{Audio NN}
The para-lingual information of vocal signals was found to convey significant information on the emotional state of a person. Even though vocal ER has been widely explored using the conventional handcrafted features, such as Mel-frequency cepstral coefficients (MFCCs) \cite{Sethu2015}, there has been a significant improvement over the recent years with the introduction of DL models. Though deep vocal ER models can be explored using spectrograms with 2D-CNNs \cite{cite6, 9607711}, as well as raw A signal with 1D-CNNs \cite{cite7}, spectrograms are found to carry significant para-lingual information pertaining to the affective state of a person \cite{Ma2018, Satt2017EfficientER}. Spectrograms have been explored with various 2D-CNNs in the literature for ER \cite{10.1145/3428690.3429153}, \cite{albanie}. Therefore, we consider spectrograms in the proposed framework along with 2D-CNN models to extract A features. In particular, Resnet18 \cite{7780459} was used for Affwild2 dataset, and the A model as shown in Table I for RECOLA dataset. Given the differences in the size of the datasets, we have used different 2D-CNN models for RECOLA and Affwild2 in order to avoid over-fitting.
(see  implementation details in Section  \ref{Training details}).
\begin{figure*}[t!]
\centering
\includegraphics[width=1.0\linewidth]{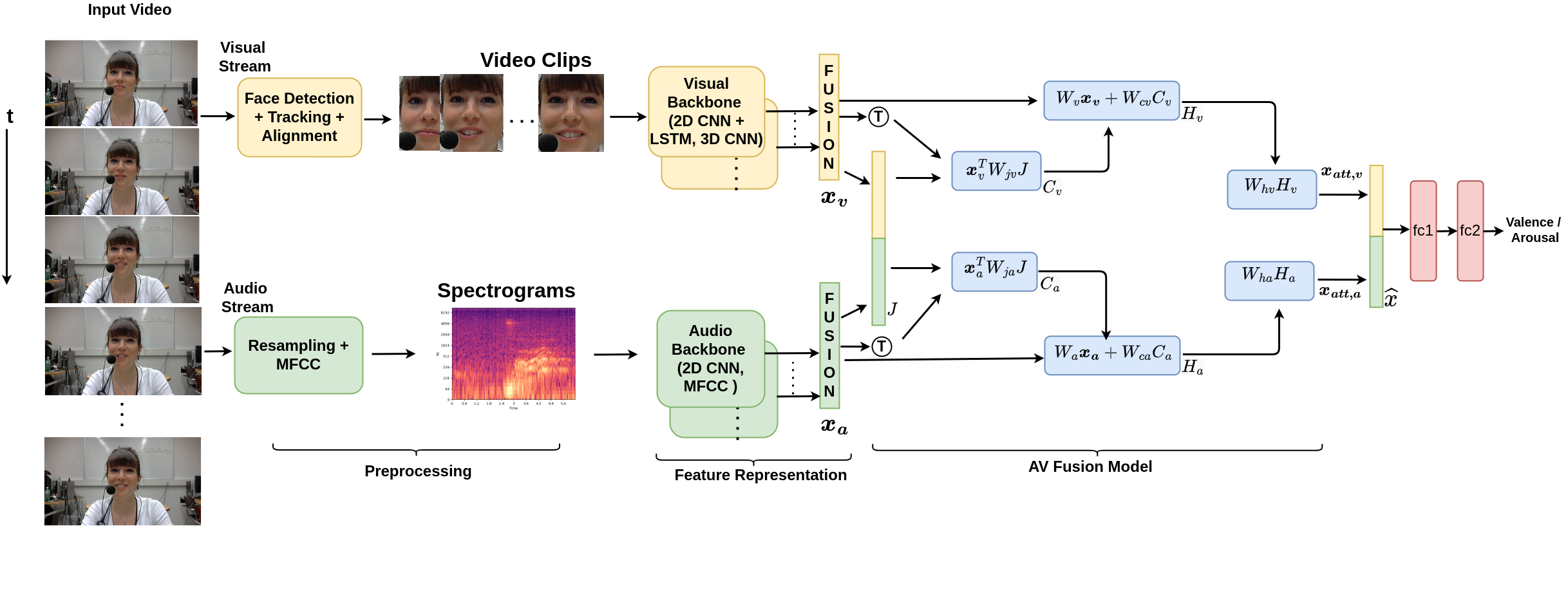}
\caption{\textbf{Joint cross-attention model proposed for A-V fusion (in testing mode).}}
\label{Block Diagram}
\end{figure*}

\subsection{Feature-Level Fusion of Multiple Backbones:}

We have also explored the fusion of features extracted from multiple backbones for both A and V modalities. Deploying multiple backbones for each modality can allow to capture diverse information for a same modality. Specifically, we have extracted V features from I3D, R3D, and 2D CNN in conjunction with Long Short Term Memory (LSTM). I3D and R3D are 3D CNN models, used to simultaneously capture the spatiotemporal relationships, which is efficient at capturing the short-term temporal relations. 2D CNN with LSTM extracts spatial features, and performs temporal modeling, which captures the long-term temporal relationships. Similarly, for A modality, combined features from a 2D CNN trained on spectrograms, and conventional handcrafted MFCC features, widely used in speech processing for many applications. 

Then we have considered two different feature-level fusion strategies to obtain a feature representation for each modality. First, we concatenate the features from all the backbones, followed by fully connected layer in order to produce a compact joint representation based on multiple diverse backbones. Feature concatenation followed by fully connected layer has been widely used in the literature for many applications. The second strategy is a more specialized feature stacking approach, where the features extracted from of multiple divers backbones and from a sequence are assembled into a block of features, and then processed by the A-V fusion model. This approach eliminates the need for training an additional fully  connected layer to combine features, as all features are trained within the fusion model.

\subsection{Joint Cross-Attentional (JCA) AV-Fusion:}
Though A-V fusion can be achieved through unified multimodal training, it was found that multimodal performance often declines over that of individual modalities \cite{9156420}. This has been attributed to a number of factors, such as differences in learning dynamics for A and V modalities \cite{9156420}, different noise topologies, with some modality streams containing more or less information for the task at hand, as well as specialised input representations \cite{Nagrani21c}. Therefore, we have trained DL models for the individual A and V modalities independently in order to extract A and V features, which is further fed to the JCA fusion model for A-V fusion that outputs final valence and arousal prediction.

For a given video sequence, the V modality carries relevant information in some video clips, whereas A modality might be more relevant for others. Since, multiple modalities convey diverse and complementary information for valence and arousal than a single modality, their  complementarity can be effectively through A and V fusion. In order to reliably fuse these modalities, we rely on cross-attention based fusion mechanism to efficiently encode the inter-modal information, while preserving the intra-modal characteristics. Though cross-attention has been conventionally applied across the features of individual modalities, we have explored cross-attention in a joint  framework. Specifically, our joint A-V feature representation is obtained by concatenating the A and V features to attend to the individual A and V features. By using the joint representation, features of each modality attend to oneself, as well as the other modality, helping to capture the semantic inter-modal relationships across A and V. The heterogeneity among the A and V modalities can also be drastically reduced by using the combined feature representation in the cross-attentional module, which further improves system performance. A block diagram of the proposed model is shown in Figure \ref{Block Diagram}.

\noindent \textbf{A) Training mode:}  
Let ${\boldsymbol X}_{\mathbf a}$ and ${\boldsymbol X}_{\mathbf v}$ represents two sets of deep feature vectors extracted for the A and V modalities, in response to a given input video sub-sequence $\boldsymbol S$ of fixed size, where 
${\boldsymbol X}_{\mathbf a}\boldsymbol =  \{ \boldsymbol x_{\mathbf a}^1, \boldsymbol x_{\mathbf a}^2, ..., \boldsymbol x_{\mathbf a}^L \boldsymbol \} \in \mathbb{R}^{d_a\times L}$ and
${\boldsymbol X}_{\mathbf v}\boldsymbol =  \{ \boldsymbol x_{\mathbf v}^1, \boldsymbol x_{\mathbf v}^2, ..., \boldsymbol x_{\mathbf v}^L \boldsymbol \} \in \mathbb{R}^{d_v\times L}$. $L$ denotes the number of non overlapping fixed-size clips sampled uniformly from $\boldsymbol S$, ${d_a}$ and ${d_v}$ represents the dimension of the A and V feature representations, respectively, and $\boldsymbol x_{\mathbf a}^{ l}$ and $\boldsymbol x_{\mathbf v}^{ l}$ denotes the A and V feature vectors, respectively, for $l = 1, 2, ..., L$ clips. Instead of applying cross-attention across the features of individual A and V modalities, we use cross-attention in a joint learning framework. The joint representation of A-V features, $\boldsymbol{J}$, is obtained by concatenating the A and V feature vectors:  
\begin{equation}
 {\boldsymbol J} = [{\boldsymbol X}_{\mathbf a} ; {\boldsymbol X}_{\mathbf v}] \in\mathbb{R}^{d\times L}
\end{equation}
where $d = {d_a} + {d_v}$ denotes the feature dimension of concatenated features.

The concatenated A-V feature representations ($\boldsymbol J$) of the given video sub-sequence ($\boldsymbol S$) is now used to attend to unimodal feature representations ${\boldsymbol X}_{\mathbf a}$ and ${\boldsymbol X}_{\mathbf v}$. The joint correlation matrix $\boldsymbol C_{\mathbf a}$ across the A features ${\boldsymbol X}_{\mathbf a}$, and the combined A-V features $\boldsymbol J$ are given by: 
\begin{equation}
   \boldsymbol C_{\mathbf a}= \tanh \left(\frac{{\boldsymbol X}_{\mathbf a}^T{\boldsymbol W}_{\mathbf j \mathbf a}{\boldsymbol J}}{\sqrt d}\right)
\end{equation}
where ${\boldsymbol W}_{\mathbf j \mathbf a} \in\mathbb{R}^{L\times L} $ represents learnable weight matrix across the A and combined A-V features, and $T$ denotes transpose operation. Similarly, the joint correlation matrix for V features are given by: 
\begin{equation}
   \boldsymbol C_{\mathbf v}= \tanh \left(\frac{{\boldsymbol X}_{\mathbf v}^T{\boldsymbol W}_{\mathbf j \mathbf v}{\boldsymbol J}}{\sqrt d}\right)
\end{equation}

The joint correlation matrices $\boldsymbol C_{\mathbf a}$ and $\boldsymbol C_{\mathbf v}$ for A and V modalities provide  a semantic measure of relevance not only across the modalities but also within the same modality. Higher correlation coefficient of the joint correlation matrices $\boldsymbol C_{\mathbf a}$ and $\boldsymbol C_{\mathbf v}$ shows that the corresponding samples are strongly correlated within the same modality as well as other modality. 
Therefore, the proposed approach is able to efficiently leverage the complementary nature of A and V modalities (i.e., inter-modal relationship) as well as intra-modal relationships, thereby improving the performance of the system. After computing the joint correlation matrices, the attention weights of A and V modalities are estimated. 

Since the dimensions of joint correlation matrices ($\mathbb{R}^{d_a\times d}$) and the features of corresponding modality ($\mathbb{R}^{L\times d_a}$) differ, we rely on a different learnable weight matrices corresponding to features of the individual modalities, and the corresponding joint correlation matrices, in order to compute attention weights of the modalities. For the A modality, the joint correlation matrix $\boldsymbol C_{\mathbf a}$ and the corresponding A features ${\boldsymbol X}_{\mathbf a}$ are combined using the learnable weight matrices $\boldsymbol W_{\mathbf c \mathbf a}$ and $\boldsymbol W_{\mathbf a}$ respectively to compute the attention weights of A modality, which is given by 
\begin{equation}
\boldsymbol H_{\mathbf a}=ReLu(\boldsymbol W_{\mathbf a} \boldsymbol X_{\mathbf a}\;+\; \boldsymbol W_{\mathbf c \mathbf a} {\boldsymbol C}_{\mathbf a}^T)
\end{equation}
where ${\boldsymbol W}_{\mathbf c \mathbf a} \in\mathbb{R}^{k\times d} $, ${\boldsymbol W}_{\mathbf a} \in\mathbb{R}^{k\times L}$ and ${\boldsymbol H}_{\mathbf a}$ represents the attention maps of the A modality. Similarly, the attention maps ($\boldsymbol H_{\mathbf v}$) of V modality are obtained as 
\begin{equation}
\boldsymbol H_{\mathbf v}=ReLu(\boldsymbol W_{\mathbf v} \boldsymbol X_{\mathbf v}\;+\; \boldsymbol W_{\mathbf c \mathbf v} {\boldsymbol C}_{\mathbf v}^T)
\end{equation}
where ${\boldsymbol W}_{\mathbf c \mathbf v} \in\mathbb{R}^{k\times d} $, ${\boldsymbol W}_{\mathbf v} \in\mathbb{R}^{k\times L}$. In our experiments, we have considered $k$ to be 32.

Finally, the attention maps are used to compute the attended features of A and V modalities. These features are obtained as:  
\begin{equation}
{\boldsymbol X}_{\mathbf a \mathbf t \mathbf t, \mathbf a} = \boldsymbol W_{\mathbf h \mathbf a}  \boldsymbol H_{\mathbf a} + \boldsymbol X_{\mathbf a}
\end{equation}
\begin{equation}
{\boldsymbol X}_{\mathbf a \mathbf t \mathbf t, \mathbf v} = \boldsymbol W_{\mathbf h \mathbf v}  \boldsymbol H_{\mathbf v} + \boldsymbol X_{\mathbf v}  
\end{equation}
where $\boldsymbol W_{\mathbf h \mathbf a} \in\mathbb{R}^{k\times L}$ and $\boldsymbol W_{\mathbf h \mathbf v} \in\mathbb{R}^{k\times L}$ denote the learnable weight matrices, respectively. The attended A and V features, ${\boldsymbol X}_{\mathbf a \mathbf t \mathbf t, \mathbf a}$ and $ {\boldsymbol X}_{\mathbf a \mathbf t \mathbf t, \mathbf v}$ are further concatenated to obtain the A-V feature representation, which is given by:  
\begin{equation}
\mathbf {\widehat X} = [{\boldsymbol X}_{\mathbf a\mathbf t\mathbf t\boldsymbol,\mathbf v} ; {\boldsymbol X}_{\mathbf a\mathbf t\mathbf t\boldsymbol,\mathbf a} ]  
\end{equation}
Finally, the A-V features are fed to the fully connected layers for the predictions of valence and arousal.

The Concordance Correlation Coefficient ($\rho_c$) has been widely used in the literature to measure the level of agreement between the predictions ($x$) and ground truth ($y$) annotations for dimensional ER \cite{cite7}. Let $\mu_x$ and $\mu_y$ represents the mean of predictions and ground truth, respectively. Similarly, if $\sigma_x^2$ and $\sigma_y^2$ denotes the variance of predictions and ground truth, respectively, then $\rho_c$ between the predictions and ground truth is:
\begin{equation}
\rho_c=\frac{2\sigma_{xy}^2}{\sigma_x^2+\sigma_y^2+(\mu_x-\mu_y)^2}
\end{equation}
where $\sigma_{xy}^2$ denotes the predictions -- ground truth covariance. 
Although MSE has been widely used as a loss function for regression models, we use $\mathcal{L} = 1 - \rho_c$ since it is standard and conventional loss function in the literature of dimensional ER literature \cite{cite7}. The parameters of our A-V fusion model ($\boldsymbol W_{\mathbf c \mathbf a}$, $\boldsymbol W_{\mathbf a}$, ${\boldsymbol W}_{\mathbf c \mathbf v}$, ${\boldsymbol W}_{\mathbf v}$, $\boldsymbol W_{\mathbf h \mathbf a}$, and $\boldsymbol W_{\mathbf h \mathbf v}$) are optimized according to this loss.
\noindent \textbf{B) Test mode:}  As shown in Figure \ref{Block Diagram},  we assume that a continuous video sequence is input to our model during inference. Feature representations $\boldsymbol x_{\mathbf a}$ and $\boldsymbol x_{\mathbf v}$ are extracted by A and V backbones for successive input clips and spectrograms, and fed to the fusion model for the prediction of valence and arousal. 

\section{Experimental Methodology}

\subsection{Datasets:}

The proposed architecture is validated on REmote COLlaborative and Affective (RECOLA) \cite{6553805} and AffWild2 \cite{Kollias}. 

\noindent  \textbf{RECOLA:}
In total, this dataset consists of 9.5 hours of multimodal recordings, which is recorded by 46 French - speaking participants, performing a collaborative task during a video conference. Among the participants, 17 are French, 3 are German and 3 are Italian. The video sequences are divided into sequences of 5 minutes each, which is annotated with a regressed intensity value for every 40 msec by 6 French speaking annotators (three male and three female). The dataset is split into three partitions: train (16 subjects), validation (15 subjects) and test (15 subjects) by balancing the age and gender of the speakers. Due to the uncontrolled spontaneous nature of expressions of the subjects, the dataset has been widely used by the research community in affective computing for various challenges such as AVEC 2015 \cite{Ringeval:2015}, AVEC 2016 \cite{Valstar:2016}, etc. Most of the existing approaches in the literature, e.g., \cite{cite6,cite7}, have validated on the dataset used for AVEC 2016 \cite{Valstar:2016} challenge, which consists of 9 subjects for training, and 9 subjects for validation. Therefore, we have also validated our proposed approach on the dataset used in AVEC 2016 challenge. 

\noindent  \textbf{Affwild2:}
Affwild2 is the largest dataset in the field of affective computing, consisting of $564$ videos collected from YouTube, all captured in-the-wild \cite{Kollias}. Sixteen of these videos display two subjects, both of which have been annotated. The annotations are provided by four experts using a joystick and the final annotations are obtained as the average of the four raters. In total, there are $2,816,832$ frames with $455$ subjects, out of which $277$ are male and $178$ female. The annotations for valence and arousal are provided continuously in the range of $\lbrack-1,1\rbrack$. The dataset is split into the training, validation and test sets. The partitioning is done in a subject independent manner, so that every subject’s data will present in only one subset. The partitioning produces 341, 71, and 152 videos for the training, validation, and test sets respectively.

\subsection{Implementation Details:} \label{Training details}

\noindent  \textbf{RECOLA:} For the \textbf{V modality}, the faces are extracted and pre-processed from the video sequences of the dataset using MTCNN model \cite{7553523}, a deep cascaded multi-task framework of face detection and alignment. Faces are resized to $224\times224$ to be fed to the I3D model \cite{8099985}. In order to generate more samples, the videos of the dataset are converted to sequences of 128 frames with a subsequence length of 16, resulting in 21,284 training samples and 16,177 validation samples. I3D uses the Inception\_v1 architecture as the base model as shown in Table \ref{visual I3D network}, which is pre-trained on Kinetics-400 dataset \cite{kinetics}, and the inflated to a 3D-CNN using RECOLA videos of facial expressions. Typically, pooling operation is performed on the last convolutional layer($512\times7\times7$) to reduce the spatial dimension to size 1 ($7\rightarrow1$), however, it may leave out useful information. Therefore, scaled dot product of audio and visual features are performed to smoothly reduce the dimension of raw visual features as in \cite{9423042}. For regularizing the network, dropout is used with $p = 0.8$ on the linear layers. The initial learning rate of the network was set to be $1e-4$ and the momentum of $0.9$ is used for Stochastic Gradient Descent (SGD). Also weight decay of $5e-4$ is used. Due to the hardware limitations and memory constraints, the batch size of the network is set to be $8$. Data augmentation is performed on the training data by random cropping, which produces scale invariant model. The number of epochs is set to be 50 and early stopping is used to obtain the best network weights.
\begin{table}[h]
\begin{center}
      \caption{Deep NN (I3D) for V Model. "Conv : 64, $7\times7\times7$, $2\times2\times2$" denotes a 3D convolutional layer of 64 filters, each of kernel size $7\times7\times7$ and stride of $2\times2\times2$. "Pool : $3\times3\times3$, $1\times2\times2$" denotes kernel size of $3\times3\times3$ and stride of $1\times2\times2$. "Linear : in = 1024, out = 256" denotes linear fully connected layer of input size 1024 and output size 256. All conv. layers are followed by Rectified Linear Units (ReLu). the linear layer in block 6 relies on a sigmoid activation function.}
    \label{visual I3D network}
\begin{tabular}{|c|c|c|c|c|c|} 
	\hline
	 \textbf{Stage}  & \textbf{Layers} & \textbf{Output size}  \\
	\hline \hline
    Input & - &  3 x 8 x 224 x 224 \\
	\hline
    \multirow{2}{*}{Block 1}  & Conv : 64, 7x7x7, 2x2x2  & \multirow{2}{*}{64 x 7 x 112 x 112} \\
      & Max pool : 1x3x3, 1x2x2 & \\
	\hline
	    \multirow{2}{*}{Block 2}  & Conv : 192, 3x3x3, 1x2x2  & \multirow{2}{*}{192 x 7 x 56 x 56} \\
      & Max pool : 3x3x3, 1x2x2 & \\
      \hline
      \multirow{2}{*}{Block 3}  & 2 x Inception Module  & \multirow{2}{*}{480 x 6 x 28 x 28} \\
	  &  Max pool : 3x3x3, 1x2x2 &  \\
      \hline
            \multirow{2}{*}{Block 4}  & 5 x Inception Module  & \multirow{2}{*}{832 x 2 x 14 x 14} \\
	  &  Max pool : 3x3x3, 1x2x2 &  \\
      \hline
            \multirow{2}{*}{Block 5}  & 2 x Inception Module  & \multirow{2}{*}{1024 x 1 x1 x 1} \\
	  &  Avg pool : 2x7x7, 1x2x2 &  \\
      \hline
	Block 6  & Linear : in = 1024, out = 256 & 256x1    \\
	\hline
	Block 7 & Linear : in = 256, out = 1 & 1x1\\
	\hline
\end{tabular}
\end{center}
\end{table}
The \textbf{A network} is composed of 3 blocks of conv. layers:  the first block has conv. layer followed by max pooling layer, the second block has two conv. layers followed by max pooling layer, and the third block has two conv. layers followed by average pooling layer, which then outputs the feature vectors. Finally, the feature vectors are fed to the linear layers to obtain the final predictions. All the conv. and linear layers layers are followed by ReLu activation functions. The vocal signal is extracted from video sub-sequences, and re-sampled to 16KHz, which is further segmented to short segments. First, we split the extracted vocal signal to 5.12 sec, which corresponds to the sequence of 128 frames of the V network. Next, the spectrogram is obtained using Discrete Fourier Transform (DFT) of length 1024 for each short vocal segment of $5.12$ sec, where the window and shift length are both 40 msec to match with the granularity of annotation frequency. Following aggregation of short-time spectra, we obtain the spectrogram of $128\times129$. The spectrogram is converted to log-power-spectrum, expressed in dB. Finally, mean and variance normalization is performed on the spectrogram. Apart from mean and variance normalization, no other voice specific processing such as silence removal, noise filtering, etc are performed. These spectrograms are then fed to the deep NN described in Table \ref{speech network}.
\begin{table}[h]
\begin{center}
      \caption{Deep NN for A Model. "Conv : 64, $5\times5$, $1\times2$" denotes a convolutional layer of 64 filters, each of kernel size $5\times5$ and stride of $1\times2$. "Pool : $4\times4$, $4\times4$" denotes kernel size of $4\times4$ and stride of $4\times4$. "Linear : in = 1024, out = 256" denotes linear fully connected layer of input size 1024 and output size 256. All conv. layers are followed by batch normalization and Rectified Linear Units (ReLu). the linear layer in block 4 relies on a sigmoid activation function.}
    \label{speech network}
\begin{tabular}{|c|c|c|c|c|c|} 
	\hline
	 \textbf{Stage}  & \textbf{Layers} & \textbf{Output size}  \\
	\hline \hline
    Input & - &  1 x 128 x 129 \\
	\hline
    \multirow{2}{*}{Block 1}  & Conv : 64, 5x5, 1x2  & \multirow{2}{*}{64 x 31 x 15} \\
      & Max pool : 4x4, 4x4 & \\
	\hline
	    \multirow{3}{*}{Block 2}  & Conv : 128, 5x5, 1x2  & \multirow{3}{*}{256 x 15 x 4} \\
	    & Conv : 256, 3x3, 1x1  & \\
      & Max pool : 2x2, 2x2 & \\
      \hline
      	    \multirow{3}{*}{Block 3}  & Conv : 512, 5x5, 1x1  & \multirow{3}{*}{1024 x 1 x 1} \\
	    & Conv : 1024, 3x3, 1x1  & \\
      & Avg pool : 13x2, 1x1 & \\
      \hline
	Block 4  & Linear : in = 1024, out = 256 & 256x1    \\
	\hline
	Block 5 & Linear : in = 256, out = 1 & 1x1\\
	\hline
\end{tabular}
\end{center}
\end{table}
The A network is trained from scratch, where the initial weights of the network are initialized with values from normal distribution. The number of epochs are set to be 100, and early stopping is used. The network is optimized using SGD with momentum of $0.9$. The initial learning rate is set to be $0.001$ and batch size is fixed to be 16. Due to the limited data, the network might be prone to over-fitting. Therefore, in order to prevent the network from over-fitting, dropout is used with p = 0.5 after the last linear layer. Also weight decay of $5e-4$ is used for all the experiments.


For the \textbf{A-V fusion network}, the size of A-V features are set to be $1024$. In the joint cross-attention module, the initial weights of the cross-attention matrix is initialized with Xavier method \cite{pmlr-v9-glorot10a}, and the weights are updated using Adam optimizer. The initial learning rate is set to be $0.001$ and batch size is fixed to be $16$. Also, dropout of $0.5$ is applied on the attended A-V features and weight decay of $5e-4$ is used for all the experiments. Due to the spontaneity of the expressions, the annotations are also found to be highly stochastic in nature. Therefore, post processing steps are applied to predictions and labels. A rigorous analysis on some of the post processing steps for annotations appears in \cite{Huang:2015}. Tzirakis et al. \cite{cite7} explored a series of post processing steps for validating their architecture on the RECOLA. Inspired by their approach, we have followed similar post processing steps to validate our architecture:  (i) median filtering with the window size ranging from $0.4$sec to $20$sec; (ii) centering the predicted values by computing the bias between annotated (ground truth) values and predicted values; (iii) matching the scaling of predicted values and annotations using the ratio of standard deviation of annotated values and predicted values. (iv) time shifting the annotations forward in time with values ranging from $0.04$ to $10$sec to compensate for delay in human annotations (delay in correspondence between the annotated values and the video frames). 

\noindent  \textbf{Affwild2:} For the \textbf{V modality}, we have used the cropped and aligned images provided by the challenge organizers \cite{kollias2021analysing}. For the missing frames in the V modality, we have considered black frames (i.e., zero pixels). Faces are resized to $224$x$224$ to be fed to the I3D network. The subsequence length and the sequence length of the videos are considered to be 8 and 64 respectively, obtained by down-sampling a sequence of 256 frames by 4. Therefore, we have 8 sub-sequences in each sequence, resulting in 1,96,265 training samples and 41,740 validation samples and 92,941 test samples. Similar to RECOLA dataset, I3D model was pre-trained on Kinetics-400 dataset \cite{kinetics}, and inflated to a 3D-CNN using Affwild2 videos of facial expressions. Instead of conventional pooling layer after the last convolutional layer, we have used scaled dot product of audio and visual features similar to that of \cite{9423042}. To regularize the network, dropout is used with $p = 0.8$ on the linear layers. The initial learning rate was set to be $1e-3$, and the momentum of $0.8$ is used for SGD. Weight decay of $5e-4$ is used. Here again, the batch size of the network is set to be $8$. Data augmentation is performed on the training data by random cropping, which produces scale invariant model. The number of epochs is set to be 50 and early stopping is used to obtain the best weights of the network.

For the \textbf{A modality}, the vocal signal is extracted from the corresponding video, and re-sampled to 44100Hz, which is further segmented to short vocal segments corresponding a sub-sequence of 256 frames of the V network. The spectrogram is obtained using Discrete Fourier Transform (DFT) of length 1024 for each short   segment, where the window length is considered to be 20 msec and the hop length to be 10 msec. Following aggregation of short-time spectra, we obtain the spectrogram of 64 x 107 corresponding to each sub-sequence of the V modality. Now a normalization step is performed on the obtained spectrograms. The spectrogram is converted to log-power-spectrum, expressed in dB. Finally, mean and variance normalization is performed on the spectrogram. Now the obtained spectrograms are fed to the Resnet18 \cite{7780459} to obtain the A features. Due to the availability of the large number of samples in the Affwild2 dataset, we trained the Resnet18 model from scratch. In order to adapt to the number of channels of the spectrogram, the first conv. layer in the Resnet18 model is replaced by single channel. The network is trained with an initial learning rate of $0.001$ and weights are optimized using Adam optimizer. The batch size is considered to be 64 and early stopping is used to obtain the best model for prediction. For the \textbf{A-V fusion network}, we have used the similar training strategy as with the RECOLA dataset.

\section{Results and Discussion}

\subsection{Ablation Study:}
\begin{table}[h]
    \centering
      \caption{\textbf{Performance of our approach with various components on the RECOLA dataset. The 2D-CNN in Table \ref{speech network} is used to extract A features in all experiments.}}
    \label{RECOLA visual network results}
\begin{tabular}{|l|c|c|c|c||c|c|c|c|c|c|} 
	\hline
	 \textbf{Method: V + Fusion}  & \textbf{Valence} & \textbf{Arousal} \\
	\hline \hline
    2D-CNN + Feature Concatenation                  & 0.538 &  0.680 \\
	\hline
	2D-CNN + LSTM                                   & 0.552 & 0.697  \\
	\hline
	I3D + Feature Concatenation                     & 0.579 &  0.732 \\
	\hline
	I3D + Cross-Attention \cite{9667055}       & 0.687 &  0.831 \\
	\hline
	I3D + Joint Cross-Attention (JCA)     & 0.728 &  0.842 \\
	\hline
	I3D; R3D; 2DCNN + JCA     & 0.762 &  0.891 \\
	\hline
\end{tabular}
\end{table}
\noindent \textbf{RECOLA:}
Table \ref{RECOLA visual network results} presents the results of our ablation study on the RECOLA validation dataset. In order to analyze the performance of our joint cross-attention model for A-V fusion, we compare with various fusion strategies widely used in the literature. One of those fusion strategies is LSTM based fusion, where the A and V features are concatenated and fed to the LSTM followed by linear layers. We have extracted V features (frame-level) using VGG 2D-CNN architecture, pretrained on FER dataset similar to \cite{8914655}, and further fine-tuned on RECOLA. Initially, we compare the proposed approach without LSTM, where the A and V features are concatenated and directly fed to linear layers. LSTM model-based fusion is evaluated by feeding the concatenated features to LSTM layer followed by fully connected layers. Given the temporal modeling of the concatenated features, the fusion performance improves over the non-LSTM based fusion strategy. We also compare the performance to I3D using baseline concatenation, where the A-V features are concatenated without attention, and fed to linear layers for valence/arousal prediction (similar to that of the fusion in \cite{8914655}). We have further compared the performance improvement of joint cross-attention fusion over that of conventional CA fusion \cite{9667055}. In case of conventional CA fusion, attention weights are computed based on the cross-correlation across the A and V modalities. The attention weights encode the semantic relevance across the A and V features. However, they do not allow the features to interact within the same modality, thereby failing to capture the temporal modeling within the same modality. Though temporal modeling across the modalities captures inter-modal relationships, and can improve the state-of-art accuracy, retaining the temporal modeling within the same modality also plays a pivotal role to capture intra-modal relationships. Therefore, we have integrated the modeling within A and V modalities, along with modeling of inter-modal relationships, and further improve system performance. Since we have introduced joint feature representation in the proposed JCA fusion model, it simultaneously captures both intra- and inter-modal relationships, and thereby outperforms the conventional CA fusion in \cite{9667055}, along with most of the widely used fusion strategies.

\begin{table}[h]
    \centering
      \caption{\textbf{Performance of our approach with various components on the Affwild2 dataset. Resnet18 \cite{7780459} is used to extract A features in all experiments.}}
    \label{Affwild2 visual network results}
\begin{tabular}{|l|c|c|c|c||c|c|c|c|c|c|} 
	\hline
	 \textbf{Method: V + Fusion}  & \textbf{Valence} & \textbf{Arousal} \\
	\hline \hline
	TSAV \cite{8914655} + Feature Concatenation & 0.531 &  0.493 \\
	\hline
	TSAV \cite{8914655} + Joint Cross-Attention (Ours)  & 0.642 &  0.592 \\
	\hline
	I3D + Feature Concatenation  & 0.498 &  0.452 \\
    \hline	
	I3D + Leader-Follower Fusion \cite{cite6} & 0.592 & 0.521  \\
	\hline
	I3D + Cross-Attention \cite{9667055} & 0.541 &  0.517 \\
	\hline
	I3D + Joint Cross-Attention (Ours)  & 0.657 &  0.580 \\
	\hline 
 	I3D; R3D; 2DCNN + JCA  & 0.725 &  0.614 \\
	\hline 
\end{tabular}
\end{table}

\noindent \textbf{Affwild2:}
Table \ref{Affwild2 visual network results} presents the results of our ablation study on the Affwild2 validation dataset. The performance of out proposed JCA fusion was compared using various A and V backbones and A-V fusion strategies. Since we used I3D for the V modality, we have compared against a V backbone based on 3D (2plus1d) CNNs \cite{8914655}. First, we implemented the backbone of TSAV \cite{8914655} with simple feature concatenation, where the extracted A and V features are concatenated, and fed to fully connected layers for valence and arousal prediction. Our proposed model provides a significant improvement in the performance. We have also analyzed when our V backbone (I3D) is used with baseline feature concatenation and  leader-follower fusion based attention \cite{cite6} with our backbones, and found that there is significant improvement in the performance over that of baseline feature concatenation. We have also implemented the conventional CA fusion \cite{9667055} with I3D backbone. Although its performance improves over that of baseline feature concatenation, it shows lower performance than leader follower attention \cite{cite6}. Finally, we have compared the proposed JCA fusion with I3D, and found that it outperforms other fusion strategies in the literature on Affwild2. By allowing the features of each modality to interact with itself and other modality, we can be effectively capture the semantic relevance of intra- and inter-modal relationships of A and V modalities for dimensional ER. We can also observe that the performance of our proposed A-V fusion model with TSAV \cite{8914655} slightly outperforms that of JCA fusion with I3D. We have further validated the proposed fusion model with multiple backbones of V and A modalities and showed further improvement in the performance of the system.

We have also explored multiple backbones for A and V modalities along with the proposed fusion model and further improved the performance of the system. As discussed in Section \ref{visual NN} and \ref{Audio NN} for V and A modalities respectively, we have used I3D, R3D and 2D CNN in conjunction with LSTM to obtain spatiotemporal features for V modality. Similarly, we have used MFCCs and spectrograms with 2D CNNs for A modality. The features of multiple backbones are fused using feature concatenation followed by fully connected layer and stacking of features in order to obtain comprehensive features for both A and V modalities. Features from multiple backbones helps to obtain diverse information of each modality and thereby improves the performance of the system. The proposed AV JCA fusion model is validated with the fusion of features from multiple backbones for both A and V modalities and the results are shown in Table \ref{fusion of backbones results}. 

\begin{table}[h]
    \centering
      \caption{\textbf{Performance of the proposed AV fusion model using fusion of features from multiple backbones for A and V modalities. FC denotes fully connected layer.}}
    \label{fusion of backbones results}
\begin{tabular}{|l|c|c|c|c||c|c|c|c|c|c|} 
	\hline
	 \textbf{Dataset} & \textbf{Fusion Method}  & \textbf{Valence} & \textbf{Arousal} \\
	\hline \hline
    \multirow{2}{*}{\textbf{RECOLA}}   &   Concatenation + FC   & 0.762 &  0.891 \\
	\cline{2-4}     & Stacking     & 0.754 & 0.865  \\
	\hline
     \multirow{2}{*}{\textbf{Affwild2}}   &   Concatenation + FC   & 0.725 &  0.614 \\
	\cline{2-4}     & Stacking   & 0.712 & 0.595  \\
	\hline
\end{tabular}
\end{table}

\begin{figure}[h]
\centering
\includegraphics[width=0.5\textwidth]{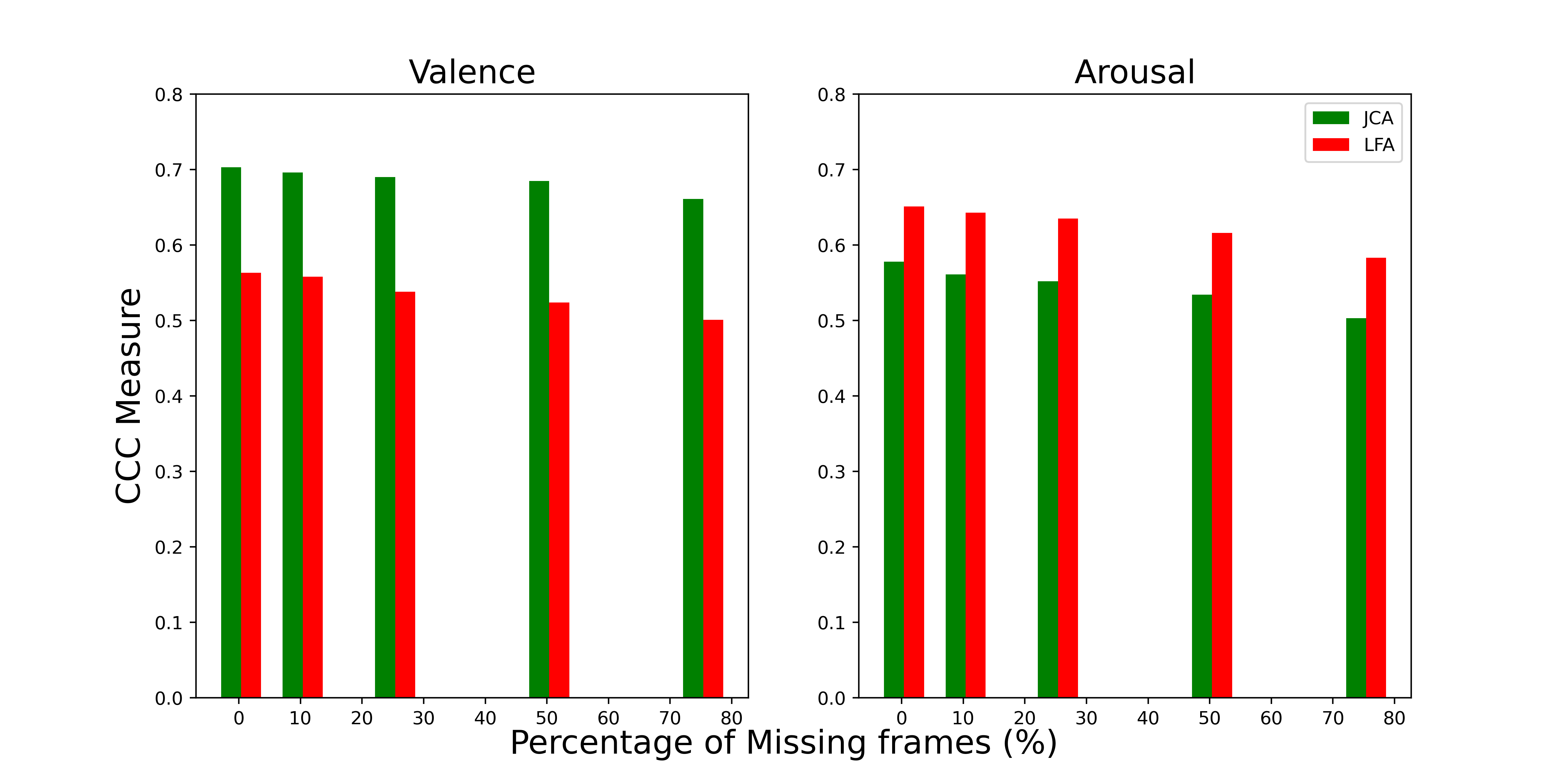}
\caption{\textbf{Performance of our proposed A-V fusion (JCA) and Leader-Follower Attention (LFA) of Zhang et al. \cite{9607460} models with a growing proportion of missing A modality.}}
\label{fig:missing audio}
\end{figure}

We have also evaluated our proposed approach for the case where a growing proportion of A is replaced by background noise in test mode. Specifically, we have randomly replaced some segments/spectrograms to reflect background noise in the video. We have tested our system on Affwild2 with a video named "16-30-1920x1080.mp4" with 5475 frames, and varied the percentage of missing spectrograms by $10$, $25$, $50$ and $100$\%. Even though spectrograms are noisy and absent, we can clearly observe that there is modest  minimal decline in CCC performance (see in Fig \ref{fig:missing audio}). In particular, since we can effectively encode the complementary relationship across modalities (by jointly modeling of intra- and inter-modal relationships), our models can sustain a high level of performance for valence. 

\begin{table*}[h]
\renewcommand{\arraystretch}{1.4}
    \centering
    \caption{ \textbf{CCC performance of proposed and state-of-art methods for A-V fusion on the RECOLA development set.  (SM represents strength modeling of SVR + BLSTM.)}}
    \label{Comparison with state-of-the-art}
    \begin{tabular}{|l||c|c|c|c|c|c||c|c|c|} 
	\hline
	 \textbf{Method -- A/V backbone}  &  \multicolumn{3}{|c|}{\textbf{Valence}} & \multicolumn{3}{|c|}{\textbf{Arousal}}  \\ \cline{2-7}
	 & \textbf{Audio}  & \textbf{Visual}  & \textbf{Fusion}  & \textbf{Audio} & \textbf{Visual} & \textbf{Fusion}\\
	 \hline
	\hline
	He et al. \cite{Helang}, AVEC 2015 -- A: LLDs; V: LLDs & 0.400 &  0.441 & 0.609 & 0.800 & 0.587 & 0.747\\
	\hline
	Han et al. \cite{Han}, IVU 2017 -- A: LLDs + SM; V: geometric features + S.M.& 0.480 &  0.592 & 0.554 & 0.760 & 0.350 & 0.685\\
	\hline
	Tzirakis et al. \cite{cite7}, JSTSP 2017 -- A: 1D-CNN; V: Resnet50 & 0.428 &  0.637 & 0.502 & 0.786 & 0.371 & 0.731 \\ \hline
	Ortega et al. \cite{8914655}, SMC 2019 -- A:LLDs; V: 2D-CNN & - & -  & 0.565 & - & - & 0.749  \\
	\hline
    Schoneval et al. \cite{cite6}, PRL 2021 -- A: Finetuned VGGish; V: Distilled CNN & 0.460 &  0.550 & 0.630 & 0.800 & 0.570 & 0.810  \\
	\hline
	 Rajasekhar et al \cite{9667055}, FG 2021 -- A: 2D-CNN; V: I3D & 0.463  & 0.642 & 0.687 & 0.822 & 0.582 & 0.831\\ 
	\hline 
	Joint Cross-Attention (Ours) -- A: 2D-CNN; V: I3D & 0.463  & 0.642 & \textbf{0.728} & 0.822 & 0.582 & \textbf{0.842}\\ 
	\hline
\end{tabular}
\end{table*}

\begin{table*}
\renewcommand{\arraystretch}{1.4}
\centering
\caption{ \textbf{CCC performance of the proposed and state-of-the-art methods for A-V fusion on the Affwild2 development set. (TCN denotes Temporal Convolutional Network.)}}
    \label{Comparison with state-of-the-art for Affwild2 validation}
    \begin{tabular}{|l||c|c|c|c|c|c||c|c|c|} 
	\hline
	 \textbf{Method} -- \textbf{A/V backbone}  &  \multicolumn{3}{|c|}{\textbf{Valence}} & \multicolumn{3}{|c|}{\textbf{Arousal}}  \\ \cline{2-7}
	 & \textbf{Audio}  & \textbf{Visual}  & \textbf{Fusion}  & \textbf{Audio} & \textbf{Visual} & \textbf{Fusion}\\
	 \hline
	\hline
    Kuhnke et al. \cite{9320301}, FGW 2020 -- A: Resnet18; V: R(2plus1)D & 0.355 & 0.463 & 0.493 & 0.359 & 0.570 & 0.613 \\
	\hline
   Zhang et al. \cite{9607460}, ICCVW 2021 -- A: VGGish; V: Resnet50 + TCN & - & 0.425 & 0.469 & - & \textbf{0.647}  & \textbf{0.649} \\
   \hline
	 Rajasekhar et al \cite{9667055}, FG 2021 -- A: Resnet18; V: I3D & 0.355  & 0.412 & 0.541 & 0.359 & 0.534 & 0.517\\ 
	\hline 
	Joint Cross-Attention (Ours) -- A: Resnet18; V: I3D & 0.355  & 0.412 & \textbf{0.657} & 0.359 & 0.534 & 0.580\\ 
	\hline
\end{tabular}
\end{table*}

\subsection{Comparison to the State-of-Art:}

\noindent \textbf{RECOLA:} 
Table \ref{Comparison with state-of-the-art} presents our comparative results against state-of-the-art A-V fusion models on the RECOLA development set. He et al. \cite{Helang} explored handcrafted LPQ-TOP features for V, and low-level descriptors (LLDs) such as MFCC, energy, etc. for A, along with physiological modalities like electro-cardiogram (ECG) and electro-dermal activity (EDA). Given the use of additional physiological modalities, and more LLD descriptors in A, as well as additional geometric features of V, the fusion performance provides significant improvement. Han et al. \cite{Han} explored LLD features for A, and facial landmark features (only geometric) for V, combined in a hierarchical fashion to leverage the individual advantages of support vector regressor (SVR) and Bidirectional Long Short-Term Memory Networks (BLSTM), and improved the performance for valence. Inspired by performance of DL models, Tzirakis et al. \cite{cite7} explored Resnet50 2D-CNN for V, and 1D-CNN on raw data for A. However, the features are directly concatenated, and fed to LSTMs. This results in a decline in CCC performance over individual modalities. The performance has been further improved by Juan et al \cite{8914655}, where they pre-train a CNN on FER for V, and LLD for A. Recently, Schoneval et al. \cite{cite6} used knowledge distillation for V, and a VGG network on  spectrograms for A. Instead of direct concatenation, they rely on two independent CNNs before concatenating them,  and showed that their fusion outperforms over individual modalities. Though deep models have improved the performance over handcrafted features, they fail to effectively leverage the complementary nature of the A-V modalities. By effectively leveraging the intra and inter-modal relationships of A and V features, the proposed model outperforms state-of-the-art approaches using joint cross-attention. 

\noindent \textbf{Affwild2:} 
Table \ref{Comparison with state-of-the-art for Affwild2 validation} shows our comparative results against relevant state-of-the-art A-V fusion models on the Affwild2 dataset. In the recent years, most of the existing work on the Affwild2 dataset have been submitted to the Affective Behavior Analysis in-the-wild (ABAW) challenges \cite{kollias2020analysing,Kollias_2021_ICCV}. Therefore, we compare our proposed approach with that of the top relevant approaches appearing in ABAW challenges for A-V fusion. 
However, the experimental protocol and training data varies widely among these approaches. We therefore re-implemented these approaches according to our experimental protocol, and analyzed the results on Affwild2 validation set for fair comparison.  Similar to our A and V backbones, Kuhnke et al \cite{9320301} also used 3D-CNNs, where R(2plus1)D model is used for visual modality and Resnet18 is used for audio modality. However, they perform simple feature concatenation without any specialized fusion model for the prediction of valence and arousal. So the fusion performance was not significantly improved over the uni-modal performance. Zhang et al \cite{9607460} explored leader follower attention model for fusion and showed minimal improvement of fusion performance over uni-modal performances. Though they have shown significant performance for arousal than valence, it is highly attributed to the visual backbone. In our proposed approach, we have shown significant improvement for fusion especially for valence than arousal. Even with vanilla CA fusion \cite{9667055}, we have shown that fusion performance for valence has been improved better than \cite{9607460} and \cite{9320301}. By deploying joint representation into the cross attententional fusion model, the fusion performance of valence has been significantly improved further. In case of arousal, though the fusion performance is lower than that of \cite{9607460} and \cite{9320301}, we can observe that it has been improved better than that of uni-modal visual performance. Therefore, the proposed approach is effective in capturing the variations spanning over a wide-range of emotions (valence) than that of the intensities of the emotions (arousal). 
\begin{table}[h]
\renewcommand{\arraystretch}{1.25}
    \centering
    \caption{ \textbf{CCC of the proposed approach compared to state-of-the-art methods for A-V fusion on Affwild2 test set.}}
    \label{Comparison with state-of-the-art for Affwild2 test}
    \begin{tabular}{|l||c|c|c|c|c|c||c|c|c|} %
	\hline
	 \textbf{Method } 
	 & \textbf{Valence}  &  \textbf{Arousal} & \textbf{Mean}\\
	 \hline  \hline
     Meng et al. \cite{9857097}  & 0.606 & 0.596 & 0.601 \\
	\hline
 	Kuhnke et al. \cite{9320301}  & 0.448 & 0.417 & 0.432  \\
	\hline
    Zhang et al. \cite{9607460} & 0.463 & 0.492 & 0.477\\
	\hline
	Wang et al. \cite{9607711} & 0.478 & 0.498 & 0.488 \\
	\hline
	Deng et al. \cite{9607738} & 0.533 & 0.454 & 0.493 \\
	\hline

	Vincent et al. \cite{9857074}  & 0.418 & 0.407 & 0.413 \\
	\hline
   JCA (Ours) & \textbf{0.451} & \textbf{0.389} &  \textbf{0.420} \\
	\hline
\end{tabular}
\end{table}

\begin{figure*}[h]
\centering
\includegraphics[width=0.8\textwidth]{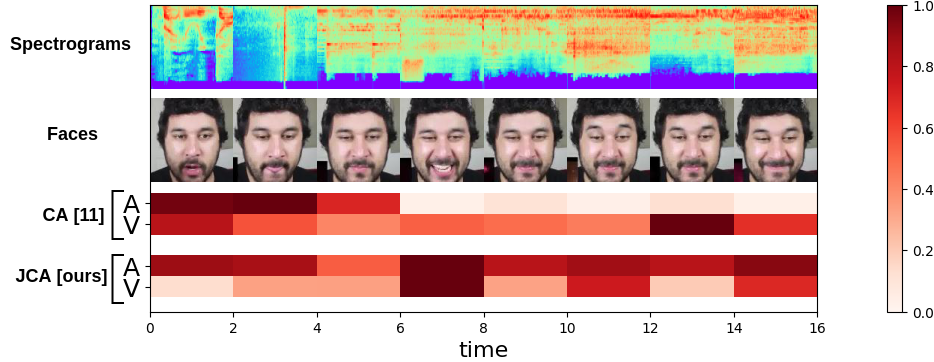}
\caption{Visualization of attention scores of our proposed A-V fusion (JCA) and CA \cite{9667055} models on video named "317" of Affwild2 dataset.}
\label{fig:vis of attention weights1}
\end{figure*}
\begin{figure*}[h]
\centering
\includegraphics[width=0.8\textwidth]{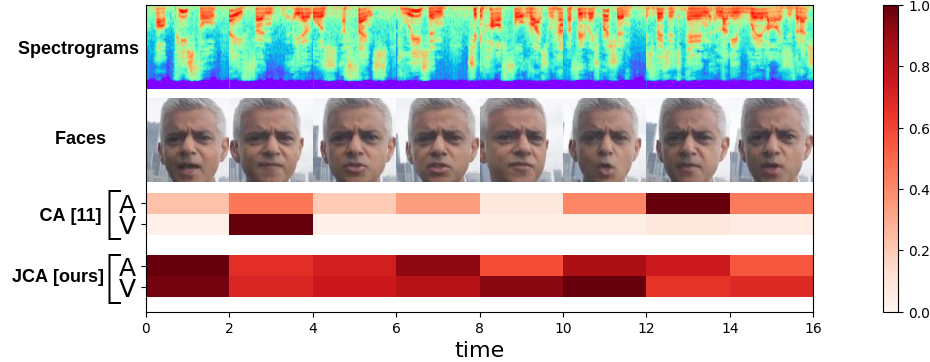}
\caption{Visualization of attention scores of our proposed A-V fusion (JCA) and CA \cite{9667055} models on video named "video92" of Affwild2 validation dataset.}
\label{fig:vis of attention weights2}
\end{figure*}
\begin{figure*}[h]
\centering
\includegraphics[width=0.8\textwidth]{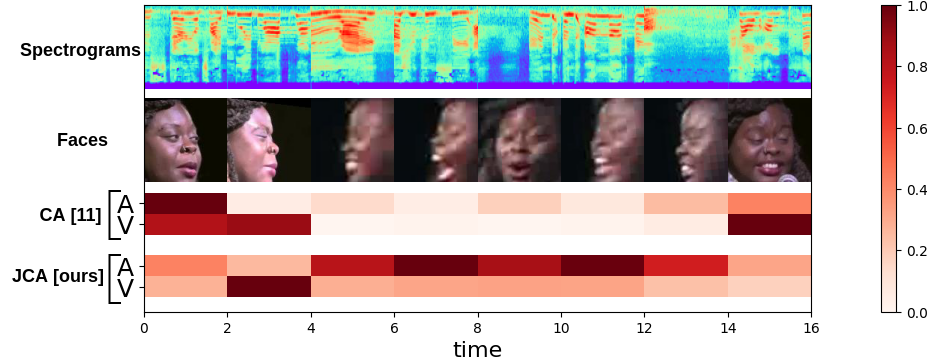}
\caption{Visualization of attention scores of our proposed A-V fusion (JCA) and CA \cite{9667055} models on video named "21-24-1920x1080" of Affwild2 validation dataset. Negative example where the proposed approach fails to focus on semantic information}
\label{fig:vis of attention weights4}
\end{figure*}
\begin{figure*}[h]
\centering
\includegraphics[width=0.8\textwidth]{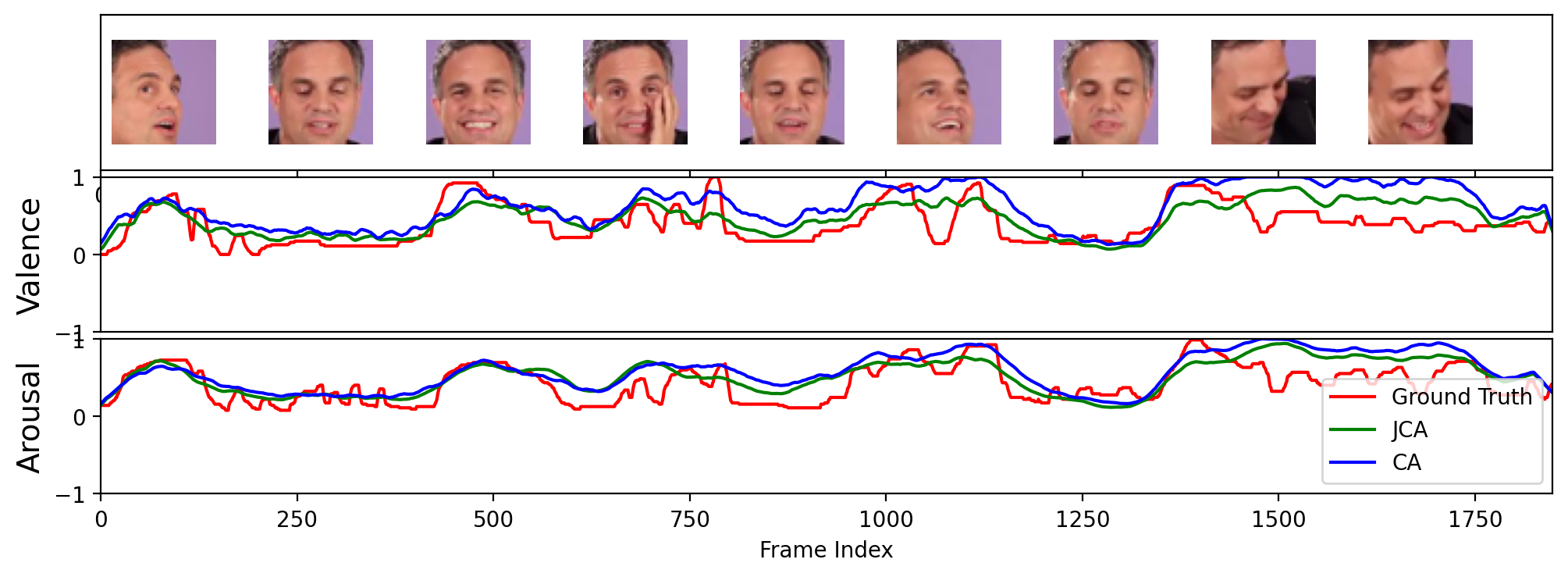}
\caption{Visualization of valence and arousal predictions over time for our proposed A-V fusion (JCA) and Cross-Attention (CA) \cite{9667055} on video named "video67" of Affwild2 validation dataset.}
\label{fig:VA}
\end{figure*}

Table \ref{Comparison with state-of-the-art for Affwild2 test} shows the results of our approach against relevant state-of-the-art A-V fusion models on the Affwild2 test set. In the recent years, several challenges such as FG2020 \cite{kollias2020analysing}, ICCV2021 \cite{Kollias_2021_ICCV} have been performed on the Affwild2 dataset as it has been the largest in-the-wild dataset in the field of affective computing. 
Kuhnke et al. \cite{8914655} proposed a two stream A-V network by using R(2plus1)D \cite{8578773} for V stream, and Resnet18 \cite{7780459} for A stream. They have also used additional masks as external inputs to guide the spatial attention of the V modality and label filtering based on multi task labels to deal with the noisy annotations of valence and arousal. Wang et al \cite{9607711} further extended their approach to perform semi-supervised learning. However they use the annotations of other ABAW challenge tasks (expression classification and action unit classification) to filter the noisy labels of valence and arousal, as well as to estimate pseudo labels for the unlabeled samples. Deng et al. \cite{9607738} proposed an iterative distillation method for modeling the uncertainty of annotations of valence and arousal and showed significant improvement in the performance. However, they have used iterative distillation of student models, which is computationally expensive as well as labels of other tasks to model the uncertainty of valence/arousal labels. Zhang et al. \cite{9607460}, Meng et al. \cite{9857097} and Vincent et al. \cite{9857074} are the only approaches, which does not use the labels of additional tasks. Meng et al. \cite{9857097} has shown significant improvement in the performance by using three external datasets along with multiple backbones of A and V modalities, whereas \cite{9607460} and \cite{9857074} uses only Affwild2 dataset similar to ours. The proposed approach performs at par with that of \cite{9607460} and better than that of \cite{9857074} in terms of valence. 
\subsection{Visual Analysis}
We have further validated the proposed approach using interpretability analysis by visualizing the attention scores of A and V modalities. In the proposed approach, we have primarily exploited the temporal attention within the same modality, as well as across the A-V modalities. So the clip-level attention scores help us to intuitively understand the important clips in the video, where the fusion attention model is focuses on the temporal sequence of A and V modalities. In order to highlight the improvement of the proposed approach w.r.t. that of the vanilla CA model \cite{9667055}, we have also plotted the attention scores of the proposed JCA model along with that of \cite{9667055}. It can be observed that the proposed JCA model is able to effectively capture the importance modalities, as well as the temporal importance within the modalities. For instance, as shown in Fig \ref{fig:vis of attention weights1}, the proposed JCA model focuses on V modality when the person smiles as the facial muscles around his nose and mouth significantly changes over time. Similarly, the proposed model assigns high attention score for A modality when there is high energy levels in the A modality. From Fig \ref{fig:vis of attention weights2}, we observe that the proposed model assigns higher attention score for clips when the person elicits knitted brows and significant facial muscle movement near his mouth, whereas \cite{9667055} fails to capture those important clips of the V modality. In both cases, we can observe that JCA assigns higher attention score to the corresponding modality when there is significant temporal variation (i.e., facial expression or tone changes), whereas the vanilla CA model \cite{9667055} fails to focus on the some of the important clips of V modality. Since the proposed JCA model leverages both the intramodal and intermodal relationships, it is able to effectively leverage the contextual information among A and V modalities. Therefore, the proposed model can efficiently exploit the importance of modalities as well as temporal importance within the modalities, resulting in better performance than that of \cite{9667055}.  

Though the proposed JCA fusion model can outperform \cite{9667055}, we observe lower performance for arousal than with valence, on both RECOLA and Affwild2 datasets. Since the proposed model considers the intra and inter variations in computing the attention scores, JCA fusion sometimes becomes mislead by assigning higher attention scores for neutral frames, and lower attention scores for more relevant clips when there is significant occlusion, blur or pose variations in the temporal sequence of the V modality. For instance, as shown in Fig \ref{fig:vis of attention weights4}, the proposed model assigns higher attention score for neutral clips, but lower attention scores for clips with more relevant facial expressions due to blur and strong pose variations. In addition to the visualization of attention scores of A and V modalities, we also visualize the valence and arousal predictions over time for videos of the Affwild2 dataset.
The proposed JCA model is able to capture the contextual relationships between A and V modalities better than that of \cite{9667055}, which helps to achieve better performance. As shown in Fig \ref{fig:VA}, we can observe that both the JCA and vanilla cross-attention models \cite{9667055} are able to track the ground truth for valence and arousal. Yet, when a fully frontal face is not available (due to pose variations), the predictions of proposed JCA model closely follow the ground truth more closely than that of \cite{9667055}, especially for valence.

\section{Conclusion}

In this paper, JCA A-V fusion model is explored for video-based dimensional ER. Contrary to the prior approaches, we leverage the intra- and inter-modal relationships across the A and V features in a unified framework. Specifically, the complementary relationship between A and V features are efficiently captured based on the correlation between the joint A-V feature representations and individual A and V features while retaining the intra-modal relationships. By jointly modeling the inter and inter-modal relationships, features of each modality attend to the other modality as well as itself, resulting in robust A and V feature representations. With the proposed model, A and V backbones are first trained individually for facial (V) and vocal (A) modalities. Then, an attention mechanism based on correlation between joint and individual features are applied to obtain the attended A and V features. Finally, the attention weighted features are concatenated, and fed to linear connected layers to predict valence and arousal values. The proposed A-V fusion model is validated experimentally on the challenging RECOLA and Affwild2 video datasets, using different A and V backbones, and different proportions of missing A segments during testing mode. Results show that the proposed model is a cost-effective approach that can outperform the state-of-the-art. It encodes inter-modal relationships, while sustaining a high level of performance, even when A segments are noisy and absent. Although the JCA AV fusion model has been proposed for dimensional emotion recognition, it can also be explored for other applications pertinent to audio visual fusion such as identity verification, event localization, etc. 

 \FloatBarrier
\bibliographystyle{IEEEtran}
\bibliography{Revised_version}

\newpage



\ifCLASSOPTIONcaptionsoff
  \newpage
\fi


\end{document}